\documentclass{article}
\usepackage{arxiv}

\usepackage[numbers,sort&compress]{natbib}
\usepackage{array}
\usepackage[caption=false,font=normalsize,labelfont=sf,textfont=sf]{subfig}
\usepackage{textcomp}
\usepackage{stfloats}
\usepackage{url}
\usepackage{verbatim}
\usepackage{graphicx}
\usepackage{balance}
\usepackage{amsmath, amsthm, amssymb}
\usepackage{bm}
\usepackage{booktabs}
\usepackage{multirow}
\usepackage{algorithm}
\usepackage{algpseudocode}

\usepackage{xcolor}

\newcommand{\safe}[2]{\textbf{#1} & \textbf{#2}}
\newcommand{\invalid}[2]{\textcolor{gray}{#1} & \textcolor{gray}{#2}}
\newcommand{\best}[2]{\textcolor{blue}{\textbf{#1}} & \textcolor{blue}{\textbf{#2}}}
\newcommand{\second}[2]{\underline{\textbf{#1}} & \underline{\textbf{#2}}}

\newtheorem{assumption}{Assumption}
\newtheorem{theorem}{Theorem}

\begin{document}

\title{Monotone Viability Learning for Offline Safe Reinforcement Learning}

\author{
Ziqian Wang
\And 
Zhen Zhang
\thanks{Corresponding author: Z. Zhang (e-mail: zzhang@tsinghua.edu.cn). The authors are both with the Department of Mechanical Engineering, Tsinghua University, Beijing 100084, China.}
}

\maketitle

\begin{abstract}
Offline safe reinforcement learning requires reliable long-horizon safety estimates from fixed data. Learning Hamilton--Jacobi (HJ) viability in this setting calls for semantic preservation, convergence, and numerical stability. To address these challenges, we propose Monotone Viability Learning (MoViL), which learns viable sets from offline transitions through monotone finite-horizon set recursion. The recursion preserves finite-horizon safety semantics and converges to the infinite-horizon viability kernel. Binary supervision maintains a fixed target scale across horizons, avoiding discount-induced numerical attenuation. We further establish sufficient conditions for recovering behavior-supported viable subsets. The learned viability estimates are used to certify actions at deployment. Dubins car experiments validate the numerical accuracy of the learned viable sets. Evaluations on safe RL benchmarks demonstrate state-of-the-art reward--cost performance. MoViL satisfies the average cost constraint on every task and achieves the lowest average cost among the compared methods. These results highlight MoViL's potential to extend long-horizon HJ viability analysis to safety-critical systems where reliable dynamics models are unavailable and online exploration is unsafe.
\end{abstract}

\section{Introduction}
\label{sec:introduction}

Safety is a central requirement for reinforcement learning (RL) in robotics and autonomous systems, where exploratory actions can cause physical damage or irreversible failures \cite{brunke2022safe}. Offline RL learns policies from previously collected data, avoiding additional environment interaction during training \cite{Prudencio2024OfflineSurvey}. Safe deployment also requires understanding how actions affect future constraint satisfaction. Many safe RL methods impose constraints on expected cumulative costs \cite{Achiam2017CPO}. These constraints limit cumulative violations but do not directly guarantee safety throughout a trajectory \cite{Zhao2023Statewise}. State-wise safety methods, including those based on control barrier functions, impose explicit conditions for safe operation \cite{Ames2019CBF,Ma2025ZeroViolation,Jha2026InputStateSafety}. Safety certification methods assess proposed actions and intervene when necessary during deployment \cite{Hsu2024SafetyFilter,Modares2024SafetyCertifier}. Learning which states admit persistent safety and which actions preserve safe continuations is therefore an important objective in offline safe RL.

Hamilton--Jacobi (HJ) reachability provides a set-based characterization of persistent safety \cite{bansal2017hamilton}. Grid-based HJ computation becomes expensive as the state dimension increases \cite{mitchell2005time}. Neural solvers improve the scalability of HJ computation using known system dynamics \cite{Bansal2021DeepReach}, while RL-based methods enable reachability-related quantities to be learned from sampled transitions \cite{fisac2019bridging,Zheng2024FISOR}. Some discounted formulations achieve contraction by modifying the original reachability semantics \cite{Akametalu2024MDR}. Other HJ formulations preserve exact semantics without establishing a corresponding contractive Bellman formulation \cite{choi2021robust}. Recent work combines semantic preservation and contraction, with a numerical implementation based on known system dynamics \cite{el2026reachability}. Neural approximation errors can compromise the reliability of learned safety boundaries \cite{Lin2024NeuralTubes,Yang2025VerifiedHJ}. This issue becomes more pronounced when discounting reduces long-horizon value magnitudes, leaving smaller margins for safety classification. Offline viability learning therefore requires a formulation that jointly addresses semantic preservation, convergence, and numerical stability.

In this article, we propose \textbf{Mo}notone \textbf{Vi}ability \textbf{L}earning (MoViL) for offline safe reinforcement learning. Building on classical viability-kernel approximation \cite{SaintPierre1994Viability}, MoViL learns long-horizon viability from offline transitions through monotone finite-horizon set recursion. At each layer, an observed transition provides a positive training example if it is violation-free and its successor belongs to the viable set estimated at the preceding layer. The learned viability information supports action-level certification during deployment, enabling a simple Gaussian   behavior-cloning policy to achieve high returns with low costs. Dubins car experiments directly validate the numerical accuracy of MoViL against an independent HJ reference. Experiments on 17 Bullet-Safety-Gym and MetaDrive tasks from DSRL \cite{Liu2024DSRL} evaluate the resulting policies across robotic control and autonomous driving settings.

The main contributions of this article are summarized as follows:
\begin{enumerate}
    \item We propose MoViL, which learns HJ viable sets from offline transitions through monotone finite-horizon recursion and uses the resulting estimates to certify policy actions.
    
    \item We prove that the finite-horizon viable sets converge to the infinite-horizon viability kernel. We also establish sufficient conditions for recovering behavior-supported viable subsets from offline data.

    \item We evaluate MoViL through numerical experiments and offline safe RL benchmarks. Dubins car experiments validate the accuracy of the learned viable sets. Across 17 offline safe RL tasks, MoViL achieves state-of-the-art reward--cost performance and the lowest average cost.
\end{enumerate}

\section{Related Work}

\subsection{Hamilton--Jacobi Reachability and Viability Learning}

Hamilton--Jacobi (HJ) reachability characterizes long-term safety through reachable sets and viability kernels \cite{bansal2017hamilton}. To connect this set-based analysis with RL, prior work has expressed safety objectives through Bellman recursions. Discounted Bellman formulations enable safety and reach-avoid values to be learned from sampled transitions \cite{fisac2019bridging,hsu2021safety}. Akametalu et al. \cite{Akametalu2024MDR} formulated reachability using minimum discounted rewards, yielding a contractive Bellman operator with guaranteed fixed-point convergence. Other formulations focus on retaining a closer connection to classical safety analysis. Control barrier-value functions connect HJ reachability with control barrier functions \cite{choi2021robust}. Travel-cost formulations derive an RL-compatible fixed point \cite{solanki2026formalizing}, while non-additive Bellman operators combine contraction with preservation of reachability semantics \cite{el2026reachability}.

These developments have enabled reachability quantities to guide safe policy learning. Reachability-constrained RL uses feasible value functions to identify viable regions \cite{yu2022reachability}. RESPO models the probability of future violations in stochastic systems \cite{Ganai2023RESPO}, while FISOR learns state and action feasibility from offline data through discounted Bellman updates and reversed expectile regression \cite{Zheng2024FISOR}. These methods make long-horizon feasibility available to policy optimization, but represent it through recursively fitted scalar values. Discounted recursions can cause value magnitudes to decrease as the horizon grows. In some formulations, the threshold used to classify feasible states does not have an exact viability interpretation. Our method addresses these issues by replacing infinite-horizon value fitting with a monotone recursion over finite-horizon viable sets. Binary membership targets retain explicit set semantics and avoid horizon-dependent target decay, while the nested recursion provides a convergent approximation to the infinite-horizon viability kernel.

\subsection{Offline Safe Reinforcement Learning}

Offline safe reinforcement learning optimizes policies from fixed datasets under safety constraints. Existing methods differ in how they incorporate safety into policy learning and execution. CPQ~\cite{xu2022constraints} uses a cost critic to screen actions during policy improvement, whereas COptiDICE~\cite{Lee2022COptiDICE} optimizes state-action occupancy subject to cost constraints. Sequence models condition action generation on return and cost signals~\cite{liu2023constrained,su2026boundary}. Generative policies can encode safety information in their latent action spaces~\cite{koirala2025latent,Zhuang2026Geometric}. V-OCBF~\cite{tayal2026vocbf} learns a control barrier function and a dynamics model from offline data to formulate a quadratic program for action filtering. During policy learning, explicit safety representations can guide policy extraction~\cite{Zheng2024FISOR} or prevent the reward critic from bootstrapping through unsafe transitions~\cite{Choudhury2026Persistent}. Our method instead uses viability estimates at deployment to certify actions proposed by a separately trained policy.
\section{Preliminaries}

\subsection{Offline Safe Reinforcement Learning}
We consider a constrained Markov decision process (CMDP)
$\mathcal{M}=(\mathcal{S},\mathcal{A},P,r,c,\rho_0,\gamma)$,
where $\mathcal{S}$ and $\mathcal{A}$ are the state and action spaces,
$P(s'|s,a)$ denotes the transition dynamics,
$r(s,a)$ and $c(s,a,s')\geq 0$ denote the reward and safety cost,
$\rho_0$ is the initial-state distribution, and
$\gamma\in(0,1)$ is the discount factor. 
We consider a state space $\mathcal{S}\subseteq\mathbb{R}^n$
and a nonempty, closed, and bounded action space
$\mathcal{A}\subseteq\mathbb{R}^m$.
For a policy $\pi$, we define
\begin{equation}
R(\pi)
=
\mathbb{E}_{\tau\sim\pi}
\left[
\sum_{t=0}^{\infty}\gamma^t r_t
\right],
\qquad
C(\pi)
=
\mathbb{E}_{\tau\sim\pi}
\left[
\sum_{t=0}^{\infty}\gamma^t c_t
\right].
\end{equation}
The safe RL objective considered in this work is
\begin{equation}
\max_{\pi}\; R(\pi)
\qquad
\mathrm{s.t.}\qquad
C(\pi)=0,
\end{equation}
which requires zero safety violation along trajectories since $c_t\geq0$.

In the offline setting, learning is based on a pre-collected dataset
\begin{equation}
\mathcal{D}
=
\{\tau_i\}_{i=1}^{N},
\qquad
\tau_i
=
\{(s_t,a_t,r_t,c_t,s_{t+1},d_t)\}_{t=0}^{T_i-1},
\end{equation}
where $d_t$ marks a true terminal transition; time-limit truncations
are recorded separately.

\begin{assumption}[Successor-state safety cost]
\label{ass:successor_state_cost}
We assume that the safety cost of a transition is determined by its successor state. Specifically, there exists
$\bar c:\mathcal{S}\rightarrow\mathbb{R}_{\geq0}$ such that
\begin{equation}
c(s,a,s')=\bar c(s').
\end{equation}
\end{assumption}

\subsection{HJ Safety Viability}

HJ reachability characterizes persistent safety through the set of states
from which safety violations can be avoided indefinitely. We study viability for the discrete-time transition system, with safety evaluated at each sampled state. For a policy $\pi$,
define
\begin{equation}
V_h^\pi(s)
=
\Pr_{\pi,P}
\left(
\exists t\geq 0:\,\bar c(s_t)>0
\,\middle|\,
s_0=s
\right),
\end{equation}
and let
\begin{equation}
V_h^*(s)=\inf_\pi V_h^\pi(s).
\end{equation}
The infinite-horizon viability kernel is
\begin{equation}
\mathcal{K}_\infty
=
\{s\in\mathcal{S}:V_h^*(s)=0\}.
\end{equation}

\section{Proposed Method}
\label{sec:method}

The proposed method combines monotone viability learning with action-level certification and policy extraction. State classifiers recursively identify finite-horizon viable regions from offline transitions. The resulting continuation labels supervise an action-conditioned safety estimator. A Gaussian behavior-cloning policy supplies nominal actions, and a cost-based fallback selects among candidate actions when the nominal action fails certification.

\subsection{Monotone Viability Formulation}
\label{subsec:viability_formulation}

\begin{figure}[!t]
\centering
\includegraphics[width=\columnwidth]{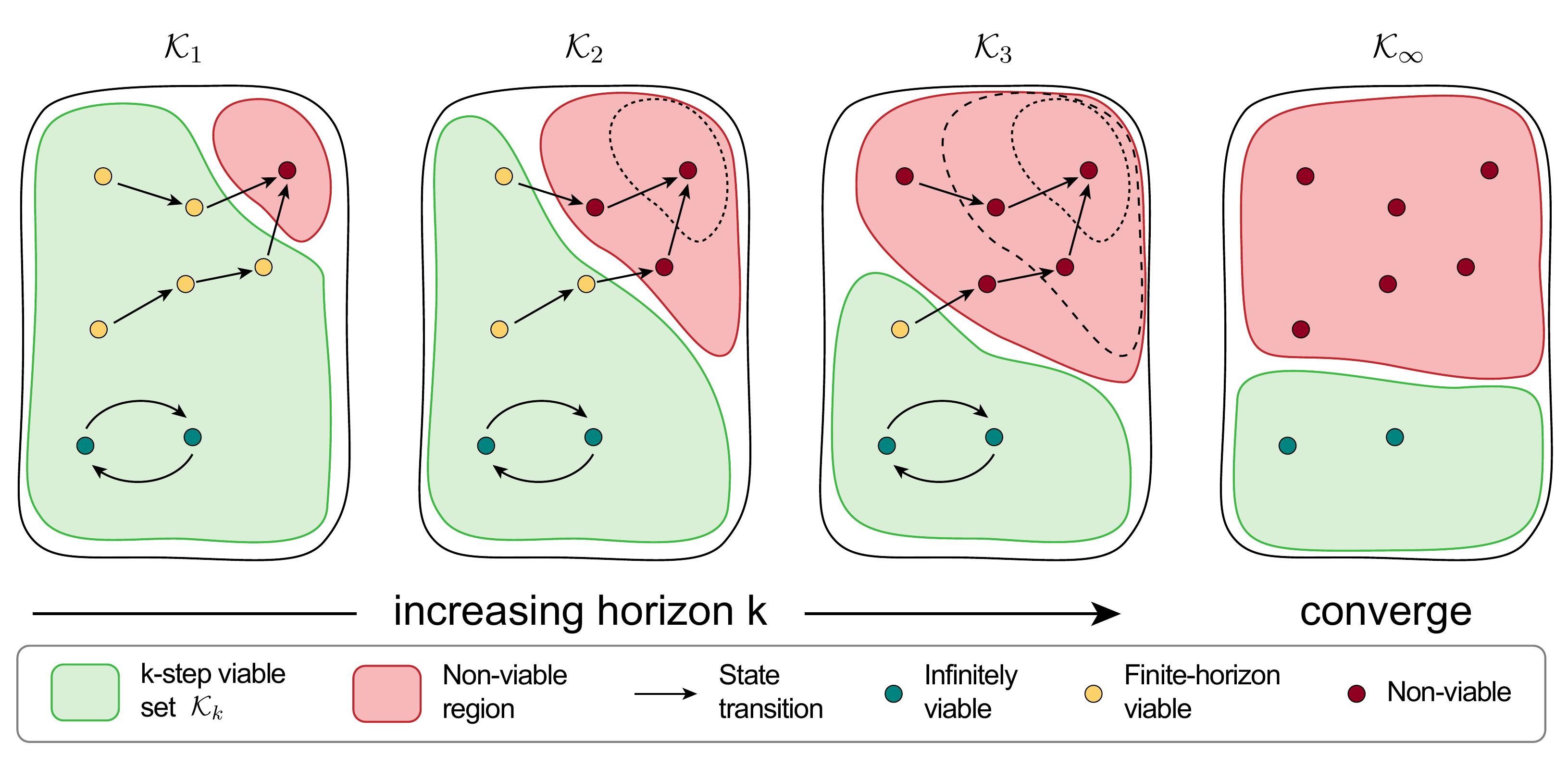}
\caption{Monotone refinement of finite-horizon viable sets. As the horizon increases, the green sets $\mathcal K_k$ shrink toward $\mathcal K_\infty$, progressively excluding states without sufficiently long safe continuations. Arrows indicate state transitions.}
\label{fig:monotone_viability_recursion}
\end{figure}

We seek convergence, semantic preservation, and numerical stability in infinite-horizon safety learning. Our approach shifts from fitting an infinite-horizon value function to recursively learning nested finite-horizon viable sets. The recursion preserves finite-horizon safety semantics and admits a monotone convergence analysis, with binary supervision maintaining a horizon-independent target scale.

Under Assumption~\ref{ass:successor_state_cost}, define the instantaneous safe set as
\begin{equation}
\mathcal K_0
=
\{s\in\mathcal S:\bar c(s)=0\}.
\label{eq:instantaneous_safe_set}
\end{equation}
Classical viability-kernel algorithms use backward set
iterations \cite{SaintPierre1994Viability,Doyen2010Stochastic}. For a measurable set $\mathcal K$, its controlled predecessor is
\begin{equation}
\operatorname{Pre}(\mathcal K)
=
\{s:\exists a\in\mathcal A,\ P(\mathcal K\mid s,a)=1\}.
\label{eq:controlled_predecessor}
\end{equation}
For deterministic transitions $s'=f(s,a)$, this condition reduces to $f(s,a)\in\mathcal K$. Starting from $\mathcal K_0$, we construct
\begin{equation}
\mathcal K_{k+1}
=
\mathcal K_k\cap\operatorname{Pre}(\mathcal K_k),
\qquad k\geq0.
\label{eq:viability_recursion}
\end{equation}
The set $\mathcal K_k$ contains initially safe states from which an admissible policy can maintain safety almost surely for the next $k$ transitions. Along this recursion, \eqref{eq:viability_recursion} is equivalent to
$\mathcal K_{k+1}=\mathcal K_0\cap\operatorname{Pre}(\mathcal K_k)$.
Fig.~\ref{fig:monotone_viability_recursion} illustrates this nested
refinement toward the infinite-horizon viability kernel.

\begin{theorem}[Convergence of monotone viability recursion]
\label{thm:viability_convergence}
Suppose $\mathcal K_0$ is closed, $\mathcal A$ is nonempty and compact, and $P$ is weakly continuous in $(s,a)$. Then the recursion in \eqref{eq:viability_recursion} satisfies
\begin{equation}
\mathcal K_{k+1}\subseteq\mathcal K_k,
\qquad
\bigcap_{k\geq0}\mathcal K_k=\mathcal K_\infty.
\label{eq:ideal_viability_limit}
\end{equation}
Moreover, $\mathcal K_\infty$ is the largest controlled invariant subset of $\mathcal K_0$ and satisfies
\begin{equation}
\mathcal K_\infty
=
\mathcal K_0\cap\operatorname{Pre}(\mathcal K_\infty).
\label{eq:viability_fixed_point}
\end{equation}
\end{theorem}

\begin{proof}
Nesting follows from \eqref{eq:viability_recursion}. For closed $\mathcal K$, weak continuity makes $(s,a)\mapsto P(\mathcal K\mid s,a)$ upper semicontinuous. Compactness of $\mathcal A$ then ensures that $\operatorname{Pre}(\mathcal K)$ is closed. Let $\mathcal K_*=\bigcap_k\mathcal K_k$. For any $s\in\mathcal K_*$, the sets
\[
\mathcal A_k(s)
=
\{a\in\mathcal A:P(\mathcal K_k\mid s,a)=1\}
\]
are nonempty, compact, and nested. Hence, an action
$a_*\in\bigcap_k\mathcal A_k(s)$ exists, and continuity of probability from above gives
\[
P(\mathcal K_*\mid s,a_*)
=
\lim_{k\to\infty}P(\mathcal K_k\mid s,a_*)=1.
\]
The correspondence $s\mapsto\bigcap_{k\geq0}\mathcal A_k(s)$ has a closed graph and nonempty compact values on $\mathcal K_*$. A measurable selector therefore defines a stationary policy that remains in $\mathcal K_*$ almost surely~\cite{BertsekasShreve1978}. Conversely, every indefinitely safe state belongs to each finite-horizon viable set. The stated assumptions ensure attainment of finite-horizon safety optima, so $V_h^*(s)=0$ also implies membership in every $\mathcal K_k$. Thus, $\mathcal K_*=\mathcal K_\infty$. Every controlled invariant subset of $\mathcal K_0$ is contained in all $\mathcal K_k$ by induction, establishing maximality. Monotonicity of $\operatorname{Pre}$ then gives \eqref{eq:viability_fixed_point}.
\end{proof}

We next approximate this set recursion from offline transitions.

\subsection{Offline Monotone Viability Learning}
\label{subsec:offline_viability}

Let $\pi_\beta(\cdot\mid s)$ denote the behavior distribution underlying the offline training dataset $\mathcal D_{\mathrm{tr}}$. State learning uses trajectory-level fitting and calibration splits, denoted by $\mathcal D_{\mathrm{fit}}$ and $\mathcal D_{\mathrm{cal}}$.
Let $(s,a,r,c,s',d)$ denote a transition from $\mathcal D_{\mathrm{tr}}$,
and let $z$ indicate a positive safety cost or a task-defined
physical-failure terminal. Time-limit truncation alone is not a
violation. For $\mathcal K\subseteq\mathcal K_0$, define
\begin{equation}
p_{\mathcal K}(s)
=
\mathbb E_{a\sim\pi_\beta(\cdot\mid s)}
\left[\Pr(z=0,\ s'\in\mathcal K\mid s,a)\right].
\label{eq:behavior_support_probability}
\end{equation}
Under deterministic transitions, $p_{\mathcal K}(s)$ is the behavior probability mass of actions reaching $\mathcal K$ without a violation. We initialize $\widehat{\mathcal K}_0=\mathcal K_0$ and write
$b_k(s)=\mathbf{1}_{\widehat{\mathcal K}_k}(s)$.

At layer $k\geq1$, we freeze the preceding membership function
$b_{k-1}$ and construct the binary target
\begin{equation}
Y_k
=
(1-z)b_{k-1}(s').
\label{eq:binary_viability_target}
\end{equation}
Safety targets also use the recorded successor membership at
nonviolating terminal or truncated records.
The conditional success probability is
\begin{equation}
p_k(s)
=
\Pr(Y_k=1\mid s)
=
p_{\widehat{\mathcal K}_{k-1}}(s).
\label{eq:layer_target_probability}
\end{equation}

Under deterministic transitions, maximizing binary successor
feasibility over actions tests whether a feasible action exists.
A positive $p_k(s)$ means that feasible actions have positive
probability mass under the behavior distribution, giving a
behavior-supported counterpart to this existence condition.

In monotone set refinement, erroneous exclusions are irreversible and reduce predecessor continuation support, potentially propagating further exclusions. Falsely retained states remain eligible for removal in later layers. This asymmetry motivates witness protection and a minimum-support threshold.

Recorded trajectories provide additional evidence of safe continuations. Let $L$ denote the number of consecutive violation-free transitions starting from the sampled record within its recorded trajectory. We define
\begin{equation}
\begin{aligned}
w_k
&=
\mathbf{1}_{\{s\in\mathcal K_0,\ L\geq k\}},\\
\widetilde Y_k
&=
\max\{Y_k,w_k\}.
\end{aligned}
\label{eq:witness_target}
\end{equation}
Let $\mathcal W_k$ consist of recorded states with $w_k=1$.
The witness sets are nested,
$\mathcal W_{k+1}\subseteq\mathcal W_k$.
Under deterministic transitions, each observed safe segment establishes finite-horizon viability, giving
$\mathcal W_k\subseteq\mathcal K_k$.

$Y_k$ uses the preceding memberships after witness protection.
In the recursion above, linked safe records with $w_k=1$ already
have $Y_k=1$, so the explicit maximum is a redundant label
safeguard retained from the implementation.
The union in \eqref{eq:learned_monotone_recursion} protects
memberships against rejection by the newly fitted network.

A state network $H_k:\mathcal S\rightarrow[0,1]$ minimizes the expectile-weighted binary cross-entropy
\begin{equation}
\begin{aligned}
\mathcal L_{H_k}
=
-\mathbb E_{\mathcal D_{\mathrm{fit}}}\bigl[
&\tau\widetilde Y_k\log H_k(s)\\
&+(1-\tau)(1-\widetilde Y_k)
\log(1-H_k(s))\bigr],
\end{aligned}
\label{eq:weighted_bce}
\end{equation}
where $\tau\in(1/2,1)$.

For a Bernoulli target with success probability $p$, the population minimizer is
\begin{equation}
h_\tau(p)
=
\frac{\tau p}
{\tau p+(1-\tau)(1-p)}.
\label{eq:binary_expectile_identity}
\end{equation}
For binary targets, this is also the population minimizer of asymmetric squared expectile loss \cite{newey1987asymmetric}, and
$\lim_{\tau\uparrow1}h_\tau(p)=\mathbf{1}_{\{p>0\}}$.
This limit connects high-expectile learning to the identification
of positively supported continuations. Let
$\widetilde p_k(s)=\Pr(\widetilde Y_k=1\mid s)$ denote the success
probability of the witness-corrected target.
It equals $p_k(s)$ whenever the safeguard leaves the conditional
labels unchanged.
For $0<\widetilde p_k(s)<1$, the optimal network output satisfies
\begin{equation}
\operatorname{logit}H_k^*(s)
=
\operatorname{logit}\widetilde p_k(s)
+
\operatorname{logit}\tau.
\label{eq:logit_probability_identity}
\end{equation}
Following the general principle of post-hoc probability
calibration \cite{Guo2017Calibration}, we fit a scalar offset $\beta_k$ by ordinary binary cross-entropy against $\widetilde Y_k$ on $\mathcal D_{\mathrm{cal}}$. The calibrated probability estimate is
\begin{equation}
\widehat p_k(s)
=
\operatorname{sigmoid}\!\left(
\operatorname{logit}H_k(s)
-
\operatorname{logit}\tau
+
\beta_k
\right).
\label{eq:probability_calibration}
\end{equation}
For a minimum positive-label probability $p_{\min}\in(0,1)$, the
equivalent network threshold is
\begin{equation}
t_k
=
\operatorname{sigmoid}\!\left(
\operatorname{logit}p_{\min}
+
\operatorname{logit}\tau
-
\beta_k
\right).
\label{eq:calibrated_threshold}
\end{equation}
$\tau$ controls the fitting emphasis on positive labels;
$p_{\min}$ controls threshold-based retention by calibrated
positive-label probability.
By construction, $H_k(s)\geq t_k$ is equivalent to
$\widehat p_k(s)\geq p_{\min}$; with exact probability recovery,
this becomes $\widetilde p_k(s)\geq p_{\min}$, independently of
$\tau$.

The membership update combines thresholding with witness protection,
\begin{equation}
\widehat{\mathcal K}_k
=
\left(
\widehat{\mathcal K}_{k-1}
\cap
\{s:\widehat p_k(s)\geq p_{\min}\}
\right)
\cup\mathcal W_k.
\label{eq:learned_monotone_recursion}
\end{equation}
The first layer preserves nesting because
$\mathcal W_1\subseteq\mathcal K_0=\widehat{\mathcal K}_0$.
For $k\geq2$,
$\mathcal W_k\subseteq\mathcal W_{k-1}
\subseteq\widehat{\mathcal K}_{k-1}$ gives the same property.
Binary membership labels $b_k$ are updated for recorded states and their successors to construct the next layer's training targets. The set inclusions in Theorem~\ref{thm:offline_viability_recovery} apply to the recursively defined sets $\widehat{\mathcal K}_k$, while $H_k$ provides a neural approximation of state viability.

\begin{theorem}[Behavior-supported viability recovery]
\label{thm:offline_viability_recovery}
Suppose the conditions of Theorem~\ref{thm:viability_convergence} hold and transitions are deterministic. Let $\mathcal C\subseteq\mathcal K_0$ be measurable and satisfy
\begin{equation}
p_{\mathcal C}(s)\geq\alpha>0,
\qquad s\in\mathcal C.
\label{eq:invariant_support_margin}
\end{equation}
Here, $\alpha$ is a uniform lower bound on the behavior probability of a violation-free transition that remains in $\mathcal C$.
Assume that each layer satisfies
\begin{equation}
\widetilde p_k(s)-\varepsilon_-
\leq
\widehat p_k(s)
\leq
\widetilde p_k(s)+\varepsilon_+,
\label{eq:asymmetric_estimation_error}
\end{equation}
for all $s\in\widehat{\mathcal K}_{k-1}$, where
$\varepsilon_-,\varepsilon_+\geq0$.
If
$\varepsilon_+<p_{\min}\leq\alpha-\varepsilon_-$,
then
\begin{equation}
\mathcal C
\subseteq
\widehat{\mathcal K}_k
\subseteq
\mathcal K_k,
\qquad k\geq0.
\label{eq:learned_set_sandwich}
\end{equation}
Consequently,
\begin{equation}
\mathcal C
\subseteq
\bigcap_{k\geq0}\widehat{\mathcal K}_k
\subseteq
\mathcal K_\infty.
\label{eq:offline_limit_sandwich}
\end{equation}
If \eqref{eq:invariant_support_margin} holds with
$\mathcal C=\mathcal K_\infty$, the limiting learned set equals
$\mathcal K_\infty$.
\end{theorem}

\begin{proof}
The claim holds at initialization. Suppose
$\mathcal C\subseteq\widehat{\mathcal K}_{k-1}
\subseteq\mathcal K_{k-1}$.
For $s\in\mathcal C$,
\[
\widehat p_k(s)
\geq
\widetilde p_k(s)-\varepsilon_-
\geq
p_{\mathcal C}(s)-\varepsilon_-
\geq p_{\min},
\]
so $s$ is retained. For
$s\in\widehat{\mathcal K}_{k-1}\setminus\mathcal K_k$,
no action enters $\widehat{\mathcal K}_{k-1}$ and no valid
$k$-step witness exists. Therefore, $\widetilde p_k(s)=0$ and
$\widehat p_k(s)\leq\varepsilon_+<p_{\min}$, excluding $s$.
Every protected witness belongs to $\mathcal K_k$.
Induction proves \eqref{eq:learned_set_sandwich}.
Taking intersections and applying
Theorem~\ref{thm:viability_convergence}
gives the limiting inclusions.
\end{proof}

The lower bound on $p_{\min}$ prevents states with
$\widetilde p_k=0$ from passing the threshold under bounded
overestimation; the upper bound retains the behavior-supported
viable subset. The interval is nonempty when
$\alpha>\varepsilon_-+\varepsilon_+$.

On fixed validation states $\mathcal S_{\mathrm{val}}$ held out from
fitting and calibration, let $u_k$ be the fraction passing all layer
thresholds $H_\ell(s)\geq t_\ell$ for $\ell=1,\ldots,k$.
Since $u_k$ is nonincreasing, its range over a window of $w\geq2$
layers is
\begin{equation}
\Delta_k^{(w)}
=
u_{k-w+1}-u_k.
\label{eq:empirical_set_change}
\end{equation}
The recursion terminates at layer $K$ once the minimum layer count
is met and this range stays within tolerance for the prescribed
number of consecutive windows; otherwise, it stops at the maximum
layer count. Fitting losses and witness-protected label retention
are monitored as diagnostics.

\subsection{Action-Level Safety Certification}
\label{subsec:action_certification}

After the state recursion terminates at layer $K$, we train two
independently initialized action-conditioned networks,
$Q_H^{(1)}$ and $Q_H^{(2)}$, using the target
\begin{equation}
Y^Q
=
(1-z)b_{K-1}(s').
\label{eq:action_safety_target}
\end{equation}
The predecessor membership $b_{K-1}$ matches the one-step
continuation condition used to train the final state layer $H_K$.
Each network minimizes ordinary binary cross-entropy over the offline training transitions,
\begin{equation}
\mathcal L_{Q_H}^{(j)}
=
\mathbb E_{\mathcal D_{\mathrm{tr}}}
\left[
\operatorname{BCE}\!\left(
Q_H^{(j)}(s,a),Y^Q
\right)
\right],
\qquad j=1,2.
\label{eq:action_safety_loss}
\end{equation}
The action-level safety score is
\begin{equation}
Q_H(s,a)
=
\min_{j\in\{1,2\}}Q_H^{(j)}(s,a).
\label{eq:twin_action_safety}
\end{equation}
An action passes certification when $Q_H(s,a)\geq t_Q$, where $t_Q$
is the action certification threshold.

The conditional success probability of $Y^Q$ is
\begin{equation}
q_K(s,a)
=
\Pr\!\left(
z=0,\ s'\in\widehat{\mathcal K}_{K-1}
\mid s,a
\right).
\label{eq:stochastic_action_probability}
\end{equation}
The uncorrected state target averages over behavior actions,
\[
p_K(s)
=
\mathbb E_{a\sim\pi_\beta(\cdot\mid s)}
[q_K(s,a)].
\]
$Q_H$ estimates action-specific success probabilities under
stochastic transitions; Theorem~\ref{thm:offline_viability_recovery}
assumes deterministic transitions.

\subsection{Certification-Guided Policy Extraction}
\label{subsec:policy_extraction}

We extract a Gaussian behavior-cloning policy from high-return, low-cost
trajectories. At execution, the nominal mean action is used if it
passes action certification. Otherwise, a \emph{cost-based fallback}
selects the action with the lowest predicted cost from a candidate
set containing the nominal action and perturbed actions.

For each training trajectory $\tau_i$, define its undiscounted return and cost,
\begin{equation}
R_i^{\mathrm{traj}}
=
\sum_{t=0}^{T_i-1}r_{i,t},
\qquad
C_i^{\mathrm{traj}}
=
\sum_{t=0}^{T_i-1}c_{i,t}.
\label{eq:trajectory_totals}
\end{equation}
Among trajectories satisfying $C_i^{\mathrm{traj}}\leq C_{\mathrm{BC}}$,
we retain the highest-return fraction $\eta_{\mathrm{BC}}$.
Their state--action pairs form the behavior-cloning dataset
$\mathcal D_{\mathrm{BC}}$.

Let $g$ denote the componentwise affine map from normalized action
coordinates $[-1,1]^m$ to the environment action bounds.
In normalized coordinates, the Gaussian policy is
\begin{equation}
\pi_{\theta,\boldsymbol\sigma}^{\mathrm{BC}}
(\widetilde a\mid s)
=
\mathcal N\!\left(
\widetilde a;
\boldsymbol\mu_\theta(s),
\operatorname{diag}(\boldsymbol\sigma^2)
\right),
\label{eq:gaussian_bc_policy}
\end{equation}
where $\boldsymbol\mu_\theta(s)$ is the state-dependent mean and
$\boldsymbol\sigma$ is a learned vector of standard deviations
shared across states. Both are trained by minimizing
\begin{equation}
\mathcal L_{\mathrm{BC}}(\theta,\boldsymbol\sigma)
=
-\mathbb E_{(s,a)\sim\mathcal D_{\mathrm{BC}}}
\left[
\log
\pi_{\theta,\boldsymbol\sigma}^{\mathrm{BC}}
\bigl(g^{-1}(a)\mid s\bigr)
\right].
\label{eq:gaussian_bc_loss}
\end{equation}

To construct candidate actions around the policy mean, let
$\epsilon_0=0$ and fix Gaussian perturbations
$\epsilon_1,\ldots,\epsilon_M\sim\mathcal N(0,I)$ across decisions.
For $j=0,\ldots,M$, define the candidate actions
\begin{equation}
a_j(s)
=
g\!\left(
\operatorname{clip}\!\left(
\boldsymbol\mu_\theta(s)
+
\boldsymbol\sigma\odot\epsilon_j,
-1,1
\right)
\right).
\label{eq:gaussian_action_candidates}
\end{equation}
The nominal action $a_0(s)$ is the clipped mean mapped to environment
coordinates.

The fallback ranks actions using a cost critic $Q_c$, trained with
a state-value network $V_c$ on $\mathcal D_{\mathrm{tr}}$ using the
common target
\begin{equation}
y^c
=
c+\gamma_c(1-d)\bar V_c(s').
\label{eq:cost_critic_target}
\end{equation}
Both networks minimize squared error to $y^c$, using $(s,a)$ for
$Q_c$ and $s$ for $V_c$. Here, $\gamma_c$ is the cost discount
factor, and $\bar V_c$ denotes the output of a slowly updated
target copy of $V_c$.
The state-value regression averages logged action outcomes at each
state, so $Q_c$ estimates the cost of taking an action and then
following the behavior policy $\pi_\beta$.

Let $j^*(s)=\operatorname*{arg\,min}_{j\in\{0,\ldots,M\}}
Q_c(s,a_j(s))$ denote the index selected by the cost critic.
The deployed policy is
\begin{equation}
\pi_{\mathrm{exec}}(s)
=
\begin{cases}
a_0(s),
&
Q_H\bigl(s,a_0(s)\bigr)\geq t_Q,
\\[1mm]
a_{j^*(s)}(s),
&
Q_H\bigl(s,a_0(s)\bigr)<t_Q,
\end{cases}
\label{eq:deployed_policy}
\end{equation}

Algorithm~\ref{alg:movil} summarizes the offline training procedure and policy execution.

\begin{algorithm}[t]
\caption{Monotone Viability Learning (MoViL)}
\label{alg:movil}
\footnotesize
\begin{algorithmic}[1]
\Require Offline data $\mathcal D_{\mathrm{tr}}$, validation states $\mathcal S_{\mathrm{val}}$, maximum depth $K_{\max}$

\Statex \textbf{Offline training}
\State Prepare $\mathcal D_{\mathrm{fit}},\mathcal D_{\mathrm{cal}}$, and trajectory witnesses; initialize $b_0=\mathbf{1}_{\mathcal K_0}$.
\For{$k=1,\ldots,K_{\max}$}
    \State Construct $Y_k=(1-z)b_{k-1}(s')$ and witness-corrected targets $\widetilde Y_k$.
    \State Train $H_k$ by weighted BCE and calibrate $\beta_k,t_k$.
    \State Update $b_k$ using \eqref{eq:learned_monotone_recursion}.
    \If{the stopping criterion on $\mathcal S_{\mathrm{val}}$ is met}
        \State \textbf{break}
    \EndIf
\EndFor
\State Set $K$ to the final recursion depth.
\State Train twin $Q_H$ with targets $Y^Q=(1-z)b_{K-1}(s')$.
\State Construct $\mathcal D_{\mathrm{BC}}$ by trajectory filtering and train Gaussian BC.
\State Train $Q_c,V_c$ on $\mathcal D_{\mathrm{tr}}$.

\Statex \textbf{Policy execution at state $s$}
\State Compute the nominal action $a_0(s)$.
\If{$Q_H(s,a_0(s))\geq t_Q$}
    \State Execute $a_0(s)$.
\Else
    \State Generate candidates $\{a_j(s)\}_{j=0}^{M}$ using fixed Gaussian perturbations.
    \State Execute $a_{j^*}(s)$, where $j^*=\arg\min_j Q_c(s,a_j(s))$.
\EndIf
\end{algorithmic}
\end{algorithm}

\section{Numerical Validation of Viability Learning}

\subsection{Experimental Setup}

\label{subsec:numerical_setup}

We consider a Dubins car navigating around a circular obstacle,
with dynamics
\begin{equation}
\dot{x}=v\cos\theta,\qquad
\dot{y}=v\sin\theta,\qquad
\dot{\theta}=u,
\label{eq:dubins_dynamics}
\end{equation}
where $(x,y)$ denotes the position, $\theta$ is the heading,
and $u\in[-1,1]$ is the steering control.
We set $v=1$ and use a control interval of $0.1$.
The obstacle is centered at the origin with radius $0.8$.
A transition is unsafe if its successor state lies inside
the obstacle.

We compare our method with FVI \cite{el2026reachability}
and FISOR \cite{Zheng2024FISOR}. FVI applies a non-additive Bellman
operator using known system dynamics, whereas FISOR learns feasibility
through discounted Bellman updates. FISOR and our method use the same
fixed offline dataset, which combines uniformly sampled and near-obstacle
initial states with steering actions sampled uniformly or at their bounds.
We compute an independent numerical HJ reference using semi-Lagrangian
recursion solely for evaluation.

\begin{table}[t]
\centering
\caption{Experimental settings for the Dubins car problem.}
\label{tab:dubins_setup}
\begin{tabular}{lc}
\toprule
\textbf{Parameter} & \textbf{Value} \\
\midrule
Offline transitions & $200,000$ \\
Maximum trajectory length & $64$ \\
HJ reference grid & $91\times91\times40$ \\
$H_k$ network & 2 hidden layers, width 192 \\
Activation function & ReLU \\
Optimizer & Adam \\
Learning rate / batch size & $3\times10^{-4}$ / $768$ \\
$\tau$ / $p_{\min}$ & $0.995$ / $0.1$ \\
Validation / evaluation states & $5,000$ / $60,000$ \\
Convergence: $w$ / tolerance / checks & $8$ / $0.005$ / $5$ \\
\bottomrule
\end{tabular}
\end{table}

The main experimental settings are summarized in Table~\ref{tab:dubins_setup}.
We evaluate viable-set approximation against the numerical HJ reference using intersection over union (IoU), false positive rate (FPR), and false negative rate (FNR), treating safe states as the positive class.
For evaluation, positions are drawn from an equal mixture of
uniform sampling over $[-3,3]^2$ and annular sampling around
the obstacle, with headings sampled uniformly from $[-\pi,\pi]$.
States are classified as safe when $H_k(s)\geq t_k$ for our method,
$V(s)\geq0$ for FVI, or $V_h(s)\leq0$ for FISOR.
The remaining hyperparameters of FVI and FISOR follow the settings
reported in their respective original papers. Quantitative results are
reported as the mean and standard deviation over three random seeds.
All experiments run on a workstation with an Intel i9-14900KF CPU
and an NVIDIA RTX 2080 Ti GPU.

\subsection{Viability Learning Results}
\label{subsec:viability_results}

\begin{figure*}[!t]
\centering
\includegraphics[width=\textwidth]{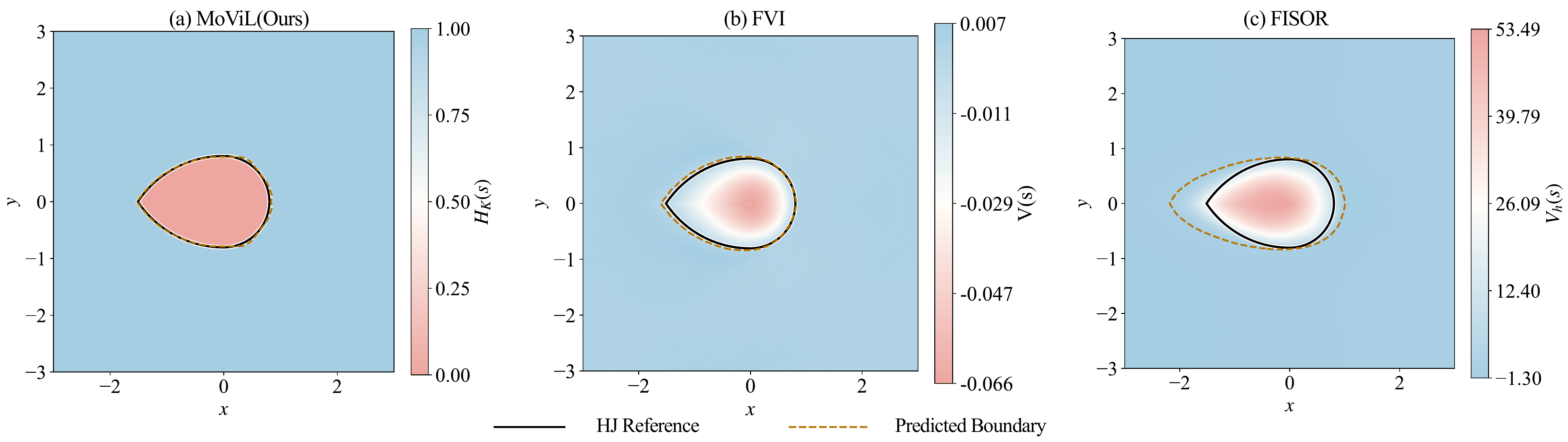}
\caption{Viability approximation on the $\theta=0$ slice of the Dubins car state space. (a) MoViL (Ours), (b) FVI, and (c) FISOR. Colors represent the learned function values. Solid and dashed lines indicate the numerical HJ reference and predicted viability boundaries, respectively.}
\label{fig:dubins_viability_comparison}
\end{figure*}

\begin{figure*}[!t]
\centering
\includegraphics[width=\textwidth]{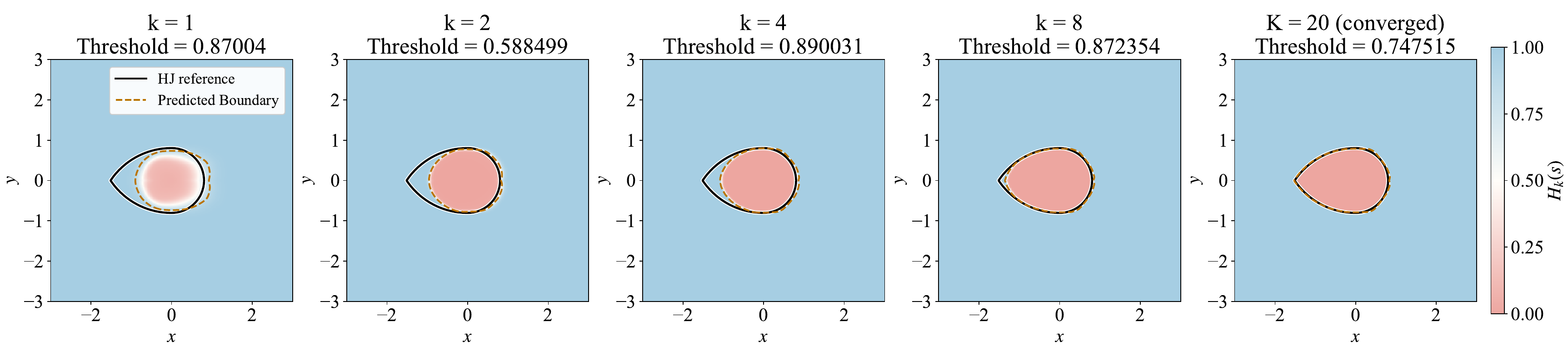}
\caption{Evolution of viability learning at layers $k=1,2,4,8,20$ on the $\theta=0$ slice. Each panel shows the current-layer prediction $H_k$ and its calibrated threshold. Solid and dashed lines indicate the numerical HJ reference and predicted boundaries, respectively.}
\label{fig:dubins_evolution}
\end{figure*}

\begin{table}[t]
\centering
\caption{Viability approximation results on the Dubins car
(mean $\pm$ standard deviation over three seeds).}
\label{tab:dubins_comparison}
\footnotesize
\setlength{\tabcolsep}{2pt}
\begin{tabular}{lcccc}
\toprule
\textbf{Method}
& \textbf{IoU (\%)} $\uparrow$
& \textbf{FPR (\%)} $\downarrow$
& \textbf{FNR (\%)} $\downarrow$
& \textbf{Time (s)} $\downarrow$ \\
\midrule
Ours
& $\mathbf{99.34\!\pm\!0.25}$
& $0.42\!\pm\!0.28$
& $\mathbf{0.62\!\pm\!0.28}$
& $\mathbf{45.94\!\pm\!2.68}$ \\
FISOR
& $94.48\!\pm\!0.23$
& $\mathbf{0.00\!\pm\!0.00}$
& $5.52\!\pm\!0.23$
& $84.40\!\pm\!1.95$ \\
FVI
& $99.07\!\pm\!0.09$
& $0.21\!\pm\!0.05$
& $0.91\!\pm\!0.09$
& $2985.38\!\pm\!5.07$ \\
\bottomrule
\end{tabular}
\end{table}

Table~\ref{tab:dubins_comparison} summarizes the viability approximation accuracy and training time of the three methods. Our method achieves the highest IoU and lowest FNR, with a slightly higher FPR than FVI. FVI attains comparable approximation accuracy but requires approximately 65 times as much training time under its original training schedule.

Fig.~\ref{fig:dubins_viability_comparison} compares the learned viability functions and predicted boundaries of the three methods with the numerical HJ reference on the $\theta=0$ slice. Fig.~\ref{fig:dubins_viability_comparison}(a) shows scores near one in viable regions and near zero in non-viable regions, with a sharp transition across the boundary. This separation reflects the binary viability classification learned by our method. The predicted boundary closely follows the numerical HJ reference. FVI also closely matches the reference boundary (Fig.~\ref{fig:dubins_viability_comparison}(b)). Its value function preserves reachability semantics through its sign \cite{el2026reachability}. Exponential discounting in its non-additive Bellman operator causes long-horizon value magnitudes to decay toward zero. The resulting small numerical margins make sign-based decisions sensitive to network fitting errors, posing a challenge as the relevant safety horizon grows. FISOR employs continuous value supervision in its discounted Bellman update, rather than binary safety labels~\cite{Zheng2024FISOR}. This method uses the zero-level set of the value function as the feasibility determination boundary, but the zero threshold does not strictly correspond to the boundary of the infinite time-domain feasible region. As shown in Fig.~\ref{fig:dubins_viability_comparison}(c), FISOR classifies a large portion of feasible regions as infeasible, exhibiting significant conservatism.

\begin{figure}[!t]
\centering
\includegraphics[width=0.8\columnwidth]{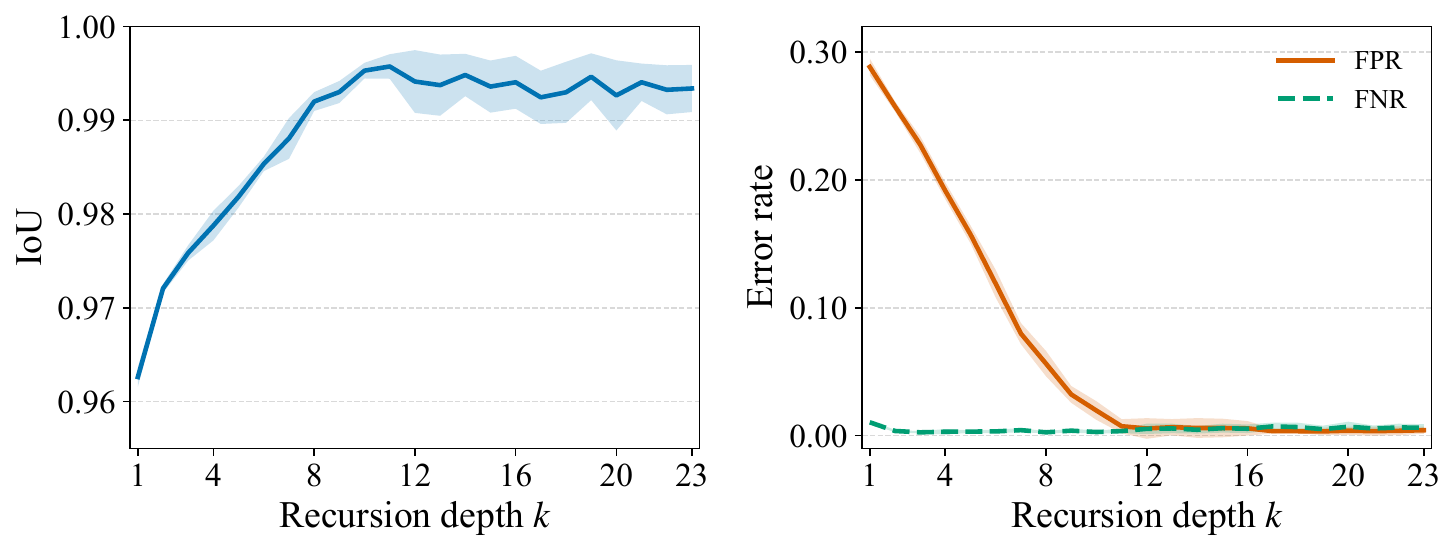}
\caption{IoU (left) and classification errors (right) across recursion layers on the Dubins car problem. Lines and shading indicate the mean and $\pm1$ standard deviation over three seeds. One run converged at $k=20$ and the other two at $k=23$. The first run's final values are reused when computing statistics at $k=21$--$23$.}
\label{fig:dubins_iou}
\end{figure}

Fig.~\ref{fig:dubins_evolution} illustrates the refinement of the learned viable set across layers. As the horizon increases, the predicted viable region progressively contracts toward the numerical HJ reference while maintaining a sharp separation between viable and non-viable regions.

Fig.~\ref{fig:dubins_iou} shows the corresponding IoU and classification errors across recursion layers. The mean IoU exceeds $0.99$ at $k=8$ and reaches a high-accuracy plateau around $k=10$. FPR decreases sharply as the recursion deepens, while FNR remains low throughout. This pattern is consistent with the design of MoViL: most reference-viable states are retained, while states without sufficiently long safe continuations are progressively excluded.

\section{Offline Safe Reinforcement Learning Experiments}

\subsection{Experimental Setup}
\label{subsec:offline_setup}

We evaluate MoViL on eight Bullet-Safety-Gym tasks~\cite{gronauer2022bullet} and nine MetaDrive tasks~\cite{li2022metadrive} from the DSRL benchmark~\cite{Liu2024DSRL}. Bullet-Safety-Gym includes four agents (Ball, Car, Drone, and Ant) performing Run and Circle tasks. Run rewards forward motion within a corridor, with safety costs for boundary violations and excessive speed. Circle rewards clockwise motion along a circular track, with costs for leaving the designated safe region. MetaDrive evaluates autonomous driving across three road complexity levels (easy, medium, hard) and three traffic densities (sparse, mean, dense), with costs for collisions and off-road driving. In both task suites, transition safety costs depend only on successor states, satisfying Assumption~\ref{ass:successor_state_cost}. We report normalized return and normalized cost, with a cost limit of $10$ for all tasks. A normalized cost no greater than $1$ indicates constraint satisfaction. Results are averaged over three random seeds, with $20$ evaluation episodes per seed.

Table~\ref{tab:offline_setup} lists the key hyperparameters for MoViL. All networks use the listed architecture, and the $Q_H$ training budget applies to each head. BC uses cosine learning-rate decay. The remaining shared hyperparameters and hardware configuration follow the Dubins car setup in Section~\ref{subsec:numerical_setup}.

\begin{table}[t]
\centering
\caption{Key hyperparameters for the offline safe RL experiments.}
\label{tab:offline_setup}
\footnotesize
\begin{tabular}{lc}
\toprule
\textbf{Parameter} & \textbf{Value} \\
\midrule
Network architecture & 2 hidden layers, width 256 \\
Batch size & $512$ \\
$H_k$ steps: first / later layers & $10,000$ / $2,000$ \\
Recursion layers: min. / max. & $20$ / $100$ \\
Convergence-validation states & $4,096$ \\
$Q_H$ / BC / cost training steps & $10,000$ / $20,000$ / $100,000$ \\
BC cost limit $C_{\mathrm{BC}}$ / fraction $\eta_{\mathrm{BC}}$ & $10$ / $0.2$ \\
Cost discount $\gamma_c$ & $0.95$ \\
Certification threshold $t_Q$ & $0.99$ \\
Action candidates $M+1$ & $10$ \\
\bottomrule
\end{tabular}
\end{table}

\subsection{Baselines}

We compare MoViL with six baselines:
(1) BCQL: A Lagrangian extension of BCQ~\cite{fujimoto2019off} for cost-constrained policy optimization.
(2) CPQ~\cite{xu2022constraints}: A Q-learning method that penalizes unsafe and out-of-distribution actions through conservative cost estimation.
(3) CDT~\cite{liu2023constrained}: A Decision Transformer conditioned on target returns and costs.
(4) FISOR~\cite{Zheng2024FISOR}: A feasibility-guided diffusion method based on HJ reachability.
(5) LSPC~\cite{koirala2025latent}: A policy optimization method with latent safety constraints learned through a conditional variational autoencoder.
(6) GSCO~\cite{Zhuang2026Geometric}: A normalizing-flow method that controls policy distribution shift through geometric constraints in latent space.

\subsection{Main Results}

\begin{table*}[t]
\centering
\caption{Performance comparison on Bullet-Safety-Gym and MetaDrive tasks.
$\uparrow$ indicates higher is better, and $\downarrow$ indicates lower is better.
\textbf{Bold}: safe policy;
\textcolor{gray}{Gray}: unsafe policy;
\textcolor{blue}{\textbf{Bold blue}}: best safe policy;
\underline{\textbf{Bold}}: second-best safe policy.
Safety is defined by normalized cost $\leq 1$, and rankings are based on reward.}
\label{tab:bullet_results}

\resizebox{\textwidth}{!}{%
\begin{tabular}{l*{14}{c}}
\toprule
\multirow{2}{*}{\textbf{Task}}
& \multicolumn{2}{c}{\textbf{BCQL}}
& \multicolumn{2}{c}{\textbf{CPQ}}
& \multicolumn{2}{c}{\textbf{CDT}}
& \multicolumn{2}{c}{\textbf{FISOR}}
& \multicolumn{2}{c}{\textbf{LSPC}}
& \multicolumn{2}{c}{\textbf{GSCO}}
& \multicolumn{2}{c}{\textbf{MoViL}} \\
\cmidrule(lr){2-3}
\cmidrule(lr){4-5}
\cmidrule(lr){6-7}
\cmidrule(lr){8-9}
\cmidrule(lr){10-11}
\cmidrule(lr){12-13}
\cmidrule(lr){14-15}
& reward $\uparrow$ & cost $\downarrow$
& reward $\uparrow$ & cost $\downarrow$
& reward $\uparrow$ & cost $\downarrow$
& reward $\uparrow$ & cost $\downarrow$
& reward $\uparrow$ & cost $\downarrow$
& reward $\uparrow$ & cost $\downarrow$
& reward $\uparrow$ & cost $\downarrow$ \\
\midrule

\multicolumn{15}{l}{\textit{Bullet-Safety-Gym}} \\

AntRun
& \invalid{0.05}{4.63}
& \safe{0.13}{0.01}
& \invalid{0.69}{1.24}
& \safe{0.45}{0.76}
& \invalid{0.94}{1.46}
& \second{0.60}{0.13}
& \best{0.61}{0.06} \\

BallRun
& \second{0.35}{0.20}
& \invalid{0.85}{13.67}
& \best{0.88}{0.86}
& \safe{0.14}{0.00}
& \safe{0.08}{0.00}
& \safe{0.21}{0.00}
& \safe{0.30}{0.00} \\

CarRun
& \invalid{0.75}{2.51}
& \safe{0.75}{0.52}
& \invalid{0.99}{1.47}
& \safe{0.80}{0.00}
& \safe{0.75}{0.22}
& \second{0.84}{0.00}
& \best{0.95}{0.00} \\

DroneRun
& \second{0.65}{0.71}
& \safe{0.26}{0.44}
& \best{0.71}{0.60}
& \safe{0.41}{0.57}
& \invalid{0.62}{1.34}
& \safe{0.55}{0.00}
& \safe{0.57}{0.00} \\

AntCircle
& \invalid{0.61}{1.42}
& \safe{0.00}{0.00}
& \invalid{0.46}{2.74}
& \safe{0.23}{0.00}
& \second{0.40}{0.78}
& \best{0.41}{0.27}
& \safe{0.34}{0.06} \\

BallCircle
& \invalid{0.79}{1.20}
& \invalid{0.40}{4.37}
& \invalid{0.79}{1.64}
& \second{0.45}{0.00}
& \invalid{0.29}{1.83}
& \safe{0.42}{0.00}
& \best{0.59}{0.01} \\

CarCircle
& \invalid{0.64}{1.80}
& \invalid{0.49}{4.48}
& \invalid{0.70}{1.20}
& \safe{0.34}{0.00}
& \safe{0.28}{0.04}
& \second{0.53}{0.00}
& \best{0.67}{0.11} \\

DroneCircle
& \invalid{0.68}{1.19}
& \invalid{-0.27}{1.29}
& \invalid{0.59}{1.56}
& \best{0.60}{0.00}
& \invalid{0.66}{1.37}
& \second{0.59}{0.10}
& \safe{0.47}{0.05} \\

\midrule
\textbf{Bullet Avg}
& \invalid{0.57}{1.71}
& \invalid{0.33}{3.10}
& \invalid{0.73}{1.41}
& \safe{0.43}{0.17}
& \safe{0.50}{0.88}
& \second{0.52}{0.07}
& \best{0.56}{0.04} \\

\midrule
\multicolumn{15}{l}{\textit{MetaDrive}} \\

Easysparse
& \invalid{0.94}{9.25}
& \safe{-0.05}{0.15}
& \safe{0.25}{0.15}
& \second{0.41}{0.50}
& \invalid{0.74}{1.55}
& \safe{0.40}{0.04}
& \best{0.56}{0.00} \\

Easymean
& \invalid{0.99}{7.22}
& \safe{-0.06}{0.00}
& \safe{0.42}{0.25}
& \safe{0.43}{0.67}
& \best{0.70}{0.68}
& \safe{0.35}{0.07}
& \second{0.55}{0.00} \\

Easydense
& \invalid{0.20}{1.76}
& \safe{-0.06}{0.16}
& \invalid{0.35}{1.17}
& \invalid{0.52}{1.26}
& \invalid{0.74}{1.48}
& \second{0.35}{0.02}
& \best{0.55}{0.02} \\

Mediumsparse
& \invalid{0.94}{2.83}
& \safe{-0.08}{0.12}
& \invalid{0.78}{1.24}
& \safe{0.43}{0.08}
& \best{0.97}{0.79}
& \safe{0.63}{0.34}
& \second{0.79}{0.04} \\

Mediummean
& \invalid{0.70}{4.45}
& \safe{-0.07}{0.16}
& \invalid{0.72}{2.74}
& \safe{0.36}{0.02}
& \best{0.92}{0.89}
& \safe{0.71}{0.54}
& \second{0.81}{0.07} \\

Mediumdense
& \invalid{0.76}{3.90}
& \safe{-0.08}{0.10}
& \invalid{0.70}{2.62}
& \safe{0.51}{0.39}
& \best{0.87}{0.88}
& \safe{0.36}{0.04}
& \second{0.76}{0.05} \\

Hardsparse
& \invalid{0.49}{3.16}
& \safe{-0.05}{0.10}
& \safe{0.26}{0.46}
& \second{0.33}{0.24}
& \invalid{0.52}{1.32}
& \safe{0.30}{0.01}
& \best{0.41}{0.04} \\

Hardmean
& \invalid{0.29}{3.80}
& \safe{-0.05}{0.15}
& \safe{0.20}{0.61}
& \safe{0.27}{0.01}
& \second{0.41}{0.57}
& \safe{0.33}{0.08}
& \best{0.41}{0.03} \\

Harddense
& \invalid{0.42}{2.95}
& \safe{-0.04}{0.12}
& \invalid{0.22}{1.38}
& \safe{0.30}{0.26}
& \invalid{0.53}{1.63}
& \second{0.36}{0.19}
& \best{0.36}{0.05} \\

\midrule
\textbf{MetaDrive Avg}
& \invalid{0.64}{4.37}
& \safe{-0.06}{0.12}
& \invalid{0.44}{1.18}
& \safe{0.40}{0.38}
& \invalid{0.71}{1.09}
& \second{0.42}{0.15}
& \best{0.58}{0.03} \\

\bottomrule
\end{tabular}%
}
\end{table*}

Table~\ref{tab:bullet_results} presents a performance comparison of the algorithms on the Bullet-Safety-Gym and MetaDrive tasks. Overall, MoViL satisfies the cost limit on every task and achieves competitive returns with the lowest average cost on both benchmarks. BCQL and CDT achieve high returns in some tasks but frequently violate safety constraints. CPQ satisfies the limit on all MetaDrive tasks, yet its uniformly negative returns indicate highly conservative behavior. FISOR satisfies the cost limit on most tasks but achieves lower average returns than MoViL. In recent strong baselines, LSPC achieves high returns in multiple tasks, but usually with higher costs. GSCO demonstrates good safety control capabilities, but its overall returns are lower than MoViL, especially on MetaDrive. In contrast, MoViL achieves higher overall returns while maintaining lower costs, demonstrating an effective balance between safety and task performance.

\subsection{Viability Certification Ablation}

\begin{figure}[!t]
\centering
\includegraphics[width=0.8\columnwidth]{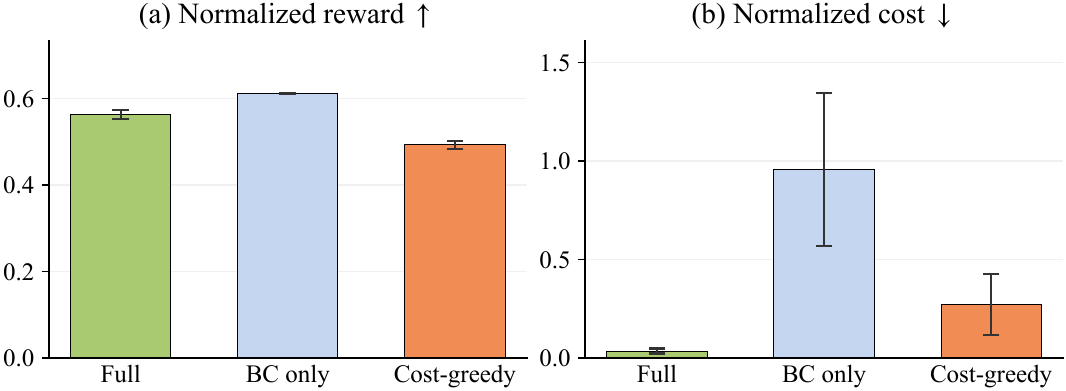}
\caption{Viability certification ablation on eight Bullet-Safety-Gym tasks. Average normalized reward and cost are reported.}
\label{fig:certification_ablation}
\end{figure}

To evaluate the contribution of viability certification, we compare the full policy with two variants, BC Only and Cost-greedy. Fig.~\ref{fig:certification_ablation} presents their average normalized reward and cost across eight Bullet-Safety-Gym tasks. BC Only directly executes the mean action of the behavior-cloning policy. It achieves the highest return, but its cost approaches the constraint limit. Cost-greedy always selects the candidate action with the lowest predicted cost. It reduces cost relative to BC Only, but results in a substantially lower return and a higher cost than the full policy. The complete policy results in the lowest average cost among the three variants, utilizing cost-based action selection only when the BC mean action does not pass certification.

\subsection{Viability Estimation Ablation}

\begin{figure}[!t]
\centering
\includegraphics[width=0.8\columnwidth]{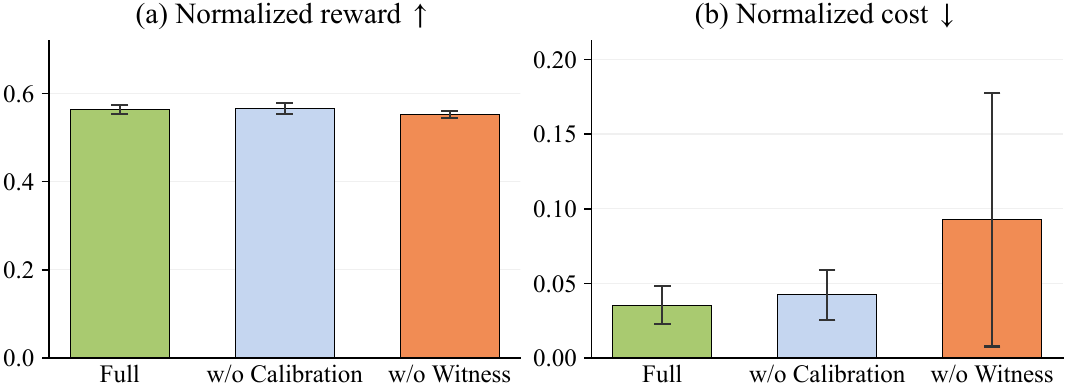}
\caption{Ablation of probability calibration and trajectory witness protection on eight Bullet-Safety-Gym tasks. Average normalized reward and cost are reported.}
\label{fig:witness_ablation}
\end{figure}

To evaluate the contributions of probability calibration and trajectory witness protection to viability learning, we compare the full method with two variants that remove each component separately. Fig.~\ref{fig:witness_ablation} presents the average normalized reward and cost across eight Bullet-Safety-Gym tasks. Removing probability calibration yields comparable reward but increases the average cost. Since calibration adjusts the threshold for retaining states during recursive learning, this result suggests that it improves the viability estimates used for subsequent action certification. Removing witness protection leads to a further increase in cost, accompanied by a reduction in reward. Witness protection preserves observed safe continuations, reducing erroneous exclusions of viable states and their propagation through subsequent training targets.

\subsection{Sensitivity to $p_{\min}$}

\begin{figure}[!t]
\centering
\includegraphics[width=0.8\columnwidth]{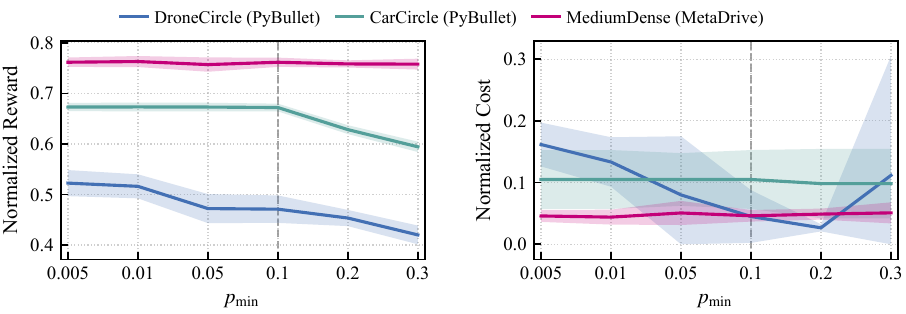}
\caption{Sensitivity to $p_{\min}$ on selected tasks.}
\label{fig:pmin_sensitivity}
\end{figure}

Fig.~\ref{fig:pmin_sensitivity} shows the sensitivity to $p_{\min}$ on two PyBullet tasks and one MetaDrive task. A larger $p_{\min}$ imposes a stricter behavior-support requirement for retaining states during viability learning. On DroneCircle, increasing $p_{\min}$ from 0.005 to 0.2 substantially reduces cost, but also lowers reward. At 0.3, cost rises again while reward decreases further. On CarCircle, reward remains stable up to $p_{\min}=0.1$, but declines at higher thresholds with little reduction in cost. In contrast, MediumDense shows only minor variations across the tested values. These results suggest that overly restrictive thresholds may exclude useful viable states without improving safety. We therefore adopt $p_{\min}=0.1$ as a reasonable balance between reward and cost.

\subsection{Sensitivity to $\tau$}

\begin{figure}[!t]
\centering
\includegraphics[width=0.8\columnwidth]{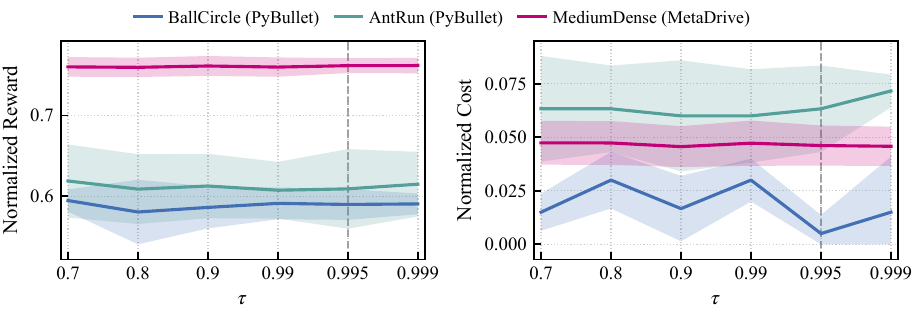}
\caption{Sensitivity to $\tau$ on selected tasks.}
\label{fig:tau_sensitivity}
\end{figure}

Fig.~\ref{fig:tau_sensitivity} shows the sensitivity to the expectile parameter $\tau$ on two PyBullet tasks and one MetaDrive task. Reward remains relatively stable across the tested values on all three tasks. The costs of AntRun and MediumDense also exhibit only minor variations. BallCircle shows larger cost fluctuations, reaching its lowest average cost at $\tau=0.995$. Increasing $\tau$ places greater emphasis on positive viability targets during training. Probability calibration compensates for the $\tau$-dependent score scaling, allowing the same behavior-support threshold to be used across different values of $\tau$. Overall, reward is relatively insensitive to $\tau$, while cost exhibits some task-dependent variations.

\begin{figure}[!t]
\centering
\includegraphics[width=\columnwidth]{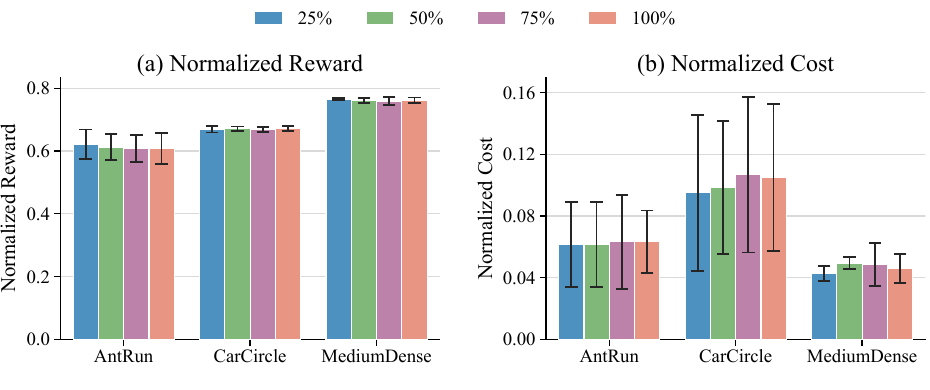}
\caption{Effect of viability training data size on policy performance across selected tasks.}
\label{fig:data_size}
\end{figure}

\subsection{Effect of Data Size}

To examine the effect of training data size on viability learning, we vary the proportion of offline data used to train $H_k$ and $Q_H$ from 25\% to 100\%, while keeping the Gaussian BC policy and cost critic fixed. Fig.~\ref{fig:data_size} compares the resulting policy performance on three representative tasks. Both reward and cost remain relatively stable across different data proportions. Even with only 25\% of the training data, MoViL achieves performance comparable to that obtained with the full dataset. These results indicate that MoViL can maintain effective viability-based certification with substantially reduced training data on the evaluated tasks.
\section{Conclusion}

We proposed MoViL for learning long-horizon HJ viability from fixed offline transitions. By approximating the infinite-horizon viability kernel through monotone finite-horizon set recursion, MoViL simultaneously achieves semantic preservation, convergence, and numerical stability. At deployment, MoViL uses the learned viability estimates to certify actions proposed by a separately trained policy. Dubins car experiments showed that the learned viable sets closely matched the numerical HJ reference. On standard safe offline RL benchmarks, MoViL achieved the lowest average normalized cost while maintaining competitive returns. Future work will investigate how viability estimates can guide the selection and composition of skills in hierarchical policies.

\bibliographystyle{unsrtnat}

\bibliography{references}



\end{document}